\documentclass[10pt,twocolumn,letterpaper]{article}

\usepackage{iccv}
\usepackage{times}
\usepackage{epsfig}
\usepackage{graphicx}
\usepackage{amsmath}
\usepackage{amssymb}
\usepackage{pdfpages}

\usepackage{pifont}
\usepackage[utf8]{inputenc} 
\usepackage[T1]{fontenc}    
\usepackage{booktabs}       
\usepackage{amsfonts}       
\usepackage{nicefrac}       
\usepackage{microtype}      
\usepackage{color}
\usepackage{bigstrut}
\usepackage{hhline}
\usepackage{colortbl}
\usepackage{tabularx}
\usepackage{subcaption}
\usepackage[numbers,sort]{natbib} 
\usepackage{arydshln}
\usepackage{array,multirow}
\usepackage{xfrac}
\usepackage{bm}
\usepackage{soul}
\usepackage[export]{adjustbox}
\usepackage{graphicx}
\usepackage{verbatim}
\usepackage{mwe}
\usepackage[shortcuts]{extdash}
\usepackage[toc,page]{appendix}
\usepackage{enumitem}
\usepackage{bbold}
\usepackage[normalem]{ulem}
\usepackage[dvipsnames]{xcolor}
\usepackage{bbm}
\usepackage{ulem}
\usepackage{float}
\usepackage{enumitem} 
\usepackage{algorithm}
\usepackage{algorithmic}

\usepackage[detect-all]{siunitx}
\usepackage{etoolbox}

\newcolumntype{L}[1]{>{\raggedright\let\newline\\\arraybackslash\hspace{0pt}}m{#1}}
\newcolumntype{R}[1]{>{\raggedleft\let\newline\\\arraybackslash\hspace{0pt}}m{#1}}
\newcolumntype{C}[1]{>{\centering\let\newline\\\arraybackslash\hspace{0pt}}m{#1}}
\newcommand{\rb}{\rotatebox{90}}%

\newcommand{\changed}[1]{{\color{blue} #1}}

\usepackage[pagebackref=true,breaklinks=true,letterpaper=true,colorlinks,bookmarks=false]{hyperref}

\iccvfinalcopy 

\def\iccvPaperID{8122} 

\ificcvfinal\fi

\begin{document}

\title{Distance-aware Quantization}

\author{Dohyung Kim
\quad\quad\quad
Junghyup Lee
\quad\quad\quad
Bumsub Ham\thanks{Corresponding author}
\\
{School of Electrical and Electronic Engineering, Yonsei University}
\\
\url{https://cvlab.yonsei.ac.kr/projects/DAQ}
}

\maketitle
\ificcvfinal\thispagestyle{empty}\fi

\begin{abstract}
We address the problem of network quantization, that is, reducing bit-widths of weights and/or activations to lighten network architectures. Quantization methods use a rounding function to map full-precision values to the nearest quantized ones, but this operation is not differentiable. There are mainly two approaches to training quantized networks with gradient-based optimizers. First, a straight-through estimator~(STE) replaces the zero derivative of the rounding with that of an identity function, which causes a gradient mismatch problem. Second, soft quantizers approximate the rounding with continuous functions at training time, and exploit the rounding for quantization at test time. This alleviates the gradient mismatch, but causes a quantizer gap problem. We alleviate both problems in a unified framework. To this end, we introduce a novel quantizer, dubbed a distance-aware quantizer (DAQ), that mainly consists of a distance-aware soft rounding (DASR) and a temperature controller. To alleviate the gradient mismatch problem, DASR approximates the discrete rounding with the kernel soft argmax, which is based on our insight that the quantization can be formulated as a distance-based assignment problem between full-precision values and quantized ones. The controller adjusts the temperature parameter in DASR adaptively according to the input, addressing the quantizer gap problem. Experimental results on standard benchmarks show that DAQ outperforms the state of the art significantly for various bit-widths without bells and whistles.
\end{abstract}



\begin{figure}[t] 
  \label{fig:teaser}
  \captionsetup{font={small}}
  \begin{center}
      \includegraphics[width=0.95\columnwidth]{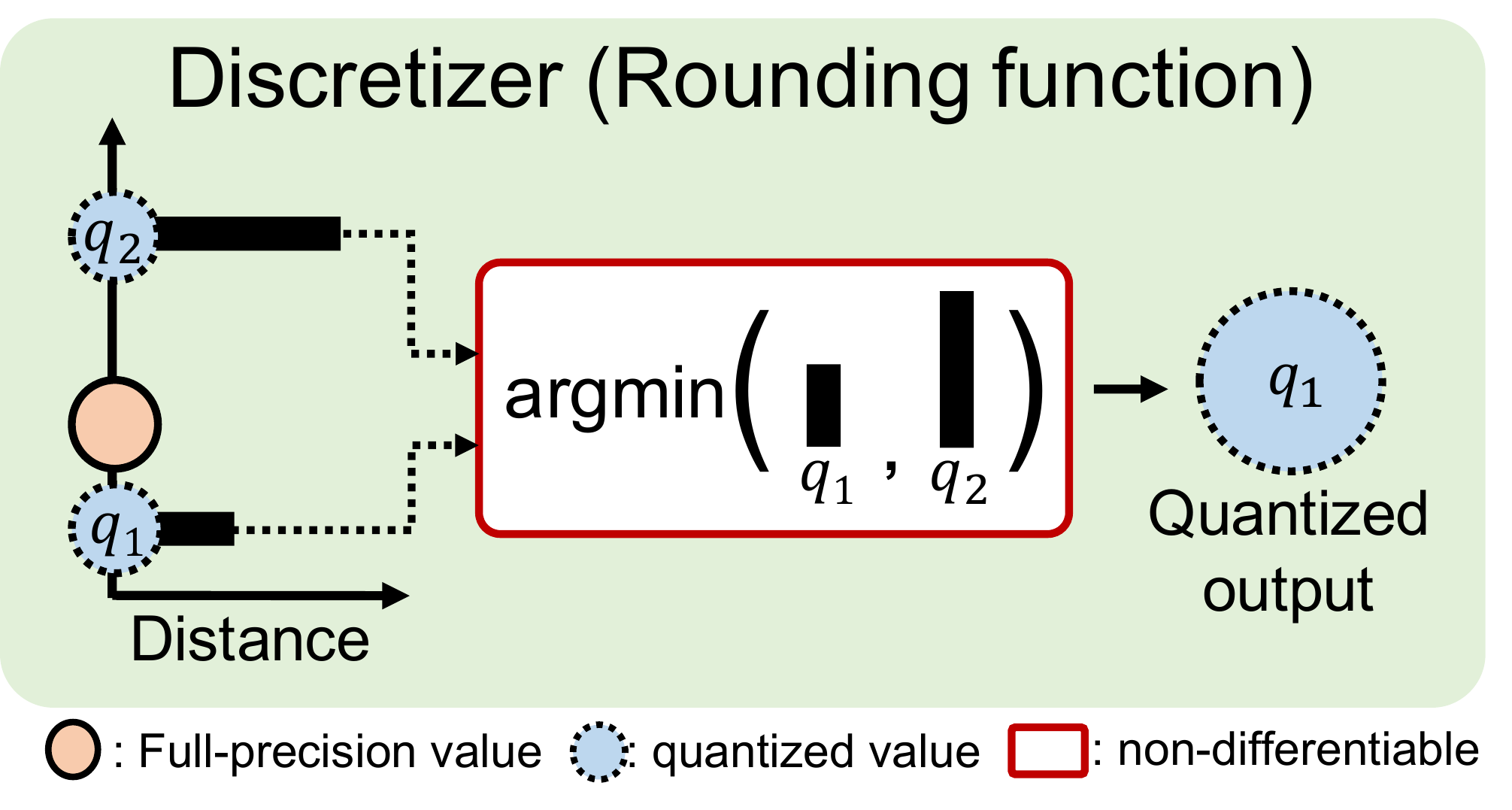}
  \end{center}
  \vspace{-0.7cm}
\caption{The discretizer takes a full-precision input, and then assigns it to the nearest quantized value,~\eg,~$q_1$ in this example. We interpret the assignment process of a discretizer as follows: It first computes the distances between the full-precision input and quantized values,~$q_1$ and $q_2$, and then applies an argmin operator over the distances to choose the quantized value. Since this operator is non-differentiable, the quantized network cannot be trained end-to-end with gradient-based optimizers. (Best viewed in color.)}
  \label{fig:teaser}
  \vspace{-0.6cm}
\end{figure}
\vspace{-0.2cm}
\section{Introduction}
\vspace{-0.2cm}
\label{sec:introduction}
Convolutional neural networks (CNNs) have made significant progress in the field of computer vision, such as image recognition~\cite{krizhevsky2012imagenet,zeiler2014visualizing}, object detection~\cite{bai2018sod,ren2015faster}, and semantic segmentation~\cite{chen2014semantic,long2015fully}. Deeper~\cite{he2016deep,xie2017aggregated} and wider~\cite{szegedy2015going} CNNs, however, require lots of parameters and FLOPs, making it difficult to deploy modern network architectures on edge devices~(\eg, mobile phones, televisions, or drones). Recent works focus on compressing networks to lighten the network architectures. Pruning~\cite{han2015deep} and distillation~\cite{hinton2015distilling} are representative techniques for network compression. The pruning removes redundant weights in a network, and the distillation encourages a compact network to have features similar to the ones obtained from a large network. The networks compressed by these techniques still exploit floating-point computations, indicating that they are not suitable for edge devices favoring fixed-point operations for power efficiency. Network quantization~\cite{rastegari2016xnor} is an alternative approach that converts full-precision weights and/or activations into low-precision ones, enabling a fixed-point inference, while reducing memory and computational cost.

Quantization methods typically use a staircase function as a quantizer, where it normalizes a full-precision value within a quantization interval, and assigns the normalized one to the nearest quantized value using a discretizer~(\ie.,~a rounding function)~\cite{jung2019learning,gong2019differentiable,esser2019learned}. Since the derivative of the rounding is zero at almost everywhere, gradient-based optimizers could not be used to train quantized networks. To address this, the straight-through estimator (STE)~\cite{bengio2013estimating} replaces the derivative of the rounding with that of identity or hard tanh functions for backward propagation. This, however, causes a gradient mismatch between forward and backward passes at training time, making the training process noisy and degrading the quantization performance at test time~\cite{esser2019learned,lin2016overcoming,yang2019quantization}. Instead of using the STE, recent methods use soft quantizers, which approximate the discrete rounding with sigmoid~\cite{yang2019quantization} or tanh~\cite{gong2019differentiable} functions, for both forward and backward passes, alleviating the gradient mismatch problem, while maintaining differentiability at training time. These approaches, on the other hand, use the discrete quantizer at inference time. That is, they exploit different quantizers (soft and discrete ones) at training and test time, resulting in a quantizer gap problem~\cite{yang2019quantization,louizos2018relaxed}. The quantizer gap might be relieved by raising a temperature parameter in the sigmoid function gradually~\cite{yang2019quantization}, such that the soft quantizer will be transformed to the discrete one eventually at training time, but this causes an unstable gradient flow.

We introduce in this paper a distance-aware quantizer (DAQ) that alleviates the gradient mismatch and quantizer gap problems in a unified framework. Our approach builds upon the insight that the discretizer (\ie.,~rounding) chooses the nearest quantized value by first computing the distances between a full-precision input and quantized values, and then applying an argmin operator over the distances w.r.t the quantized values~(Fig.~\ref{fig:teaser}). Motivated by this, we propose a distance-aware soft rounding~(DASR) that approximates the discrete rounding accurately using a kernel soft argmax~\cite{lee2019sfnet}, while maintaining differentiability, alleviating the gradient mismatch problem. We also introduce a temperature controller that adjusts a temperature parameter in DASR adaptively depending on the distances between the full-precision input and quantized values. This imposes DASR to have the same output as the discrete rounding, addressing the quantizer gap problem. We apply our DAQ to quantize weights and/or activations for various network architectures, and achieve state-of-the-art results on standard benchmarks, clearly demonstrating the effectiveness of our approach. To our knowledge, it is the first approach to alleviating both gradient mismatch and quantizer gap problems jointly. We summarize the main contributions of this paper as follows: 
\vspace{-0.2cm}
\begin{itemize}[leftmargin=*]
  \item[$\bullet$] We propose a novel differentiable approximation of the discrete rounding function, dubbed DASR, allowing to train quantization networks end-to-end, while alleviating the gradient mismatch problem. 
  \vspace{-0.2cm} 
  \item[$\bullet$] We introduce a temperature controller, which adjusts the temperature parameter in DASR adaptively, to address the quantizer gap problem.
    \vspace{-0.2cm}
  \item[$\bullet$] We set a new state of the art on standard benchmarks, and provide an extensive analysis of our approach, demonstrating the effectiveness of DAQ.
\end{itemize}
 
\begin{table}
\setlength{\tabcolsep}{0.1em}
\footnotesize
\captionsetup{font={small}}
\begin{center}
\begin{tabular}{C{1.7cm}|C{1.5cm}C{1.5cm}|C{1.5cm}C{1.5cm}}
\multirow{2}{*}{Method} & \multicolumn{2}{c|}{\bf{Gradient mismatch}} & \multicolumn{2}{c}{\bf{Quantizer gap}}    \\
& Forward & Backward & Training & Test  \\
\hline\hline
STE-based~\cite{bengio2013estimating} & rounding & identity & rounding & rounding \\
QNet~\cite{yang2019quantization} & sigmoid & sigmoid & sigmoid & rounding \\
Ours & DASR w/ $\beta^*$ & DASR w/ $\beta^*$ & DASR w/ $\beta^*$ & rounding \\
\hline
\end{tabular}
\vspace{-2mm}
\caption{Comparison of quantization methods for gradient mismatch (left) and quantizer gap (right) problems. We denote by~$\beta^*$ a temperature parameter adjusted by our temperature controller. We exploit the same DASR with~$\beta^*$ in both forward and backward passes, alleviating the gradient mismatch problem. In addition, adjusting the temperature enables DASR, used in the training stage, to have the same output as the discrete rounding at test time, addressing the quantizer gap problem.}
\vspace{-1cm}
\label{tab:comparison}
\end{center}
\end{table}

\vspace{-0.3cm}
\section{Related work}
\vspace{-0.2cm}
\label{sec:related}

\paragraph{Network quantization.}
\label{sec:related_quantization}

Early works on network quantization focus on discretizing network weights alone into binary~\cite{courbariaux2015binaryconnect,lin2015neural,hou2016loss}, ternary~\cite{li2016ternary,zhu2016trained}, or multi bits~\cite{hou2018loss,zhou2017incremental}, which however still requires lots of computational cost to process full-precision activations. Later, both weights and activations are quantized~\cite{courbariaux2016binarized,rastegari2016xnor,zhou2016dorefa}, achieving a better compromise in terms of memory and accuracy. Recent approaches exploit non-uniform quantization levels~\cite{cai2017deep,zhang2018lq} or transition points~\cite{park2018value} to minimize quantization errors. Other methods propose to learn quantization intervals~\cite{jung2019learning,gong2019differentiable,esser2019learned} or clipping ranges of activations~\cite{choi2018pact}. These approaches use a staircase function as a quantizer with a discretizer being a rounding function. Since the gradients of the rounding function are zero at almost everywhere, training quantization networks suffers from a vanishing gradient problem. The STE avoids this problem by replacing the zero derivative of the rounding function with that of a continuous one~\cite{bengio2013estimating}. More specifically, it exploits the rounding in a forward pass, while using~\eg.,~identity or tanh functions, in a backward pass. Although the STE allows to train quantized networks with gradient-based optimizers, the gradient mismatch between forward and backward passes degrades the quantization performance drastically. In contrast, we exploit the same quantizer in both forward and backward passes, alleviating the gradient mismatch problem (Table.~\ref{tab:comparison}(left)).

\begin{figure*}[t] 
  \label{fig:overview}
  \captionsetup{font={small}}
  \begin{center}
      \includegraphics[width=0.95\textwidth]{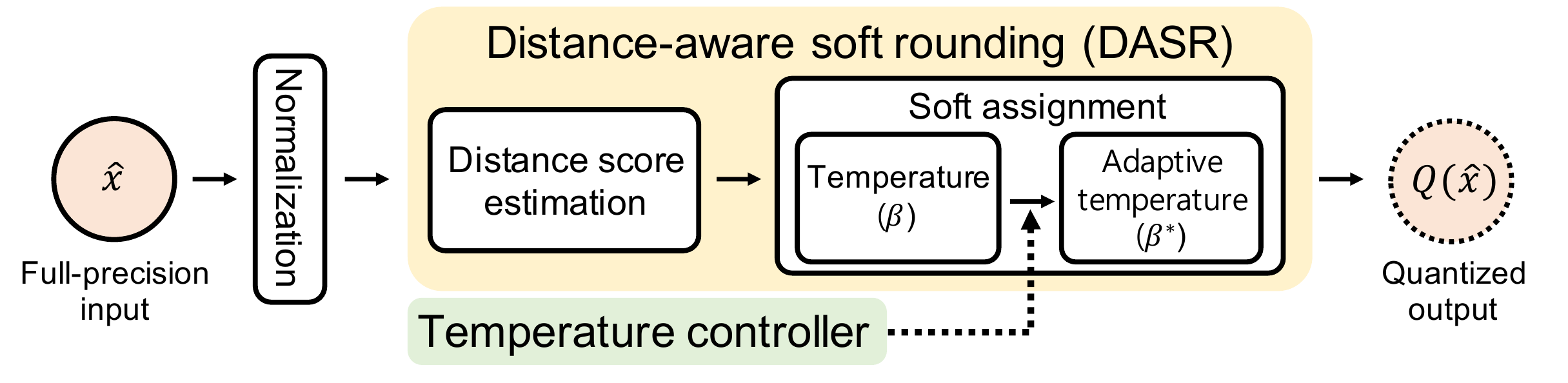}
  \end{center}
  \vspace{-0.5cm}
  \caption{{\bf{An overview of DAQ.}} Our quantizer $Q$ mainly consists of DASR with a temperature controller. DAQ first normalizes a full-precision input~$\hat{x}$. DASR inputs the normalized input, and computes distance scores w.r.t quantized values. It then assigns the input to the nearest quantized value~$Q(\hat{x})$. For the assignment, we exploit a differentiable version of the argmax with an adaptive temperature~$\beta^*$, obtained from our controller.}
  \vspace{-0.6cm}
  \label{fig:overview}
\end{figure*}

Soft quantizers approximate the discrete rounding with sigmoid~\cite{yang2019quantization} or tanh~\cite{gong2019differentiable} functions. This allows to use continuous functions in both forward and backward passes at training time, alleviating the gradient mismatch problem. These approaches, however, use the rounding function for quantization at inference time, which causes a quantizer gap problem. That is, the outputs of soft and discrete quantizers are different, degrading the quantization performance significantly. To avoid the quantizer gap problem, QNet~\cite{yang2019quantization} transforms the soft quantizer towards the discrete one gradually at training time by raising a temperature parameter in a sigmoid function. The gradients of the soft quantizer, however, vanish or explode, as the parameter increases, causing an unstable gradient flow. DSQ~\cite{gong2019differentiable} incorporates the soft quantizer with the STE. Specifically, it uses the discrete quantizer as in the STE for a forward pass, while exploiting the derivative of the soft quantizer in a backward pass. This alleviates the quantizer gap problem at inference, but the use of STE causes the gradient mismatch. Our method is similar to soft quantizers in that both approximate the discrete one, while maintaining differentiability, alleviating the gradient mismatch problem. On the contrary, we do not exploit continuous functions such as sigmoid or tanh for the approximation, and also address the quantizer gap problem in a unified framework by adjusting a temperature parameter adaptively (Table.~\ref{tab:comparison}(right)). 

Closely related to our work, RQ~\cite{louizos2018relaxed} defines categorical distributions over quantization grids, and samples a quantized value with the Gumbel soft argmax operator~\cite{jang2016categorical}. SHVQ~\cite{agustsson2017soft} designs a codebook which is a finite set of vectors (\ie.,~codewords) for vector quantization, and then assigns vectors,~\eg.,~network parameters, to one of the codewords using the soft argmax~\cite{chapelle2010gradient}. They~\cite{louizos2018relaxed,agustsson2017soft}, however, also suffer from the quantizer gap problem, since the outputs of soft and discrete quantizers are not the same. In addition, RQ performs a stochastic quantization during training, while quantizing networks in a deterministic manner at inference. This mismatch between training and inference stages degrades the performance significantly~\cite{krishnamoorthi2018quantizing}. We avoid this problem by designing DASR using a kernel soft argmax~\cite{lee2019sfnet}, quantizing networks deterministically in both training and inference stages.

\vspace{-0.4cm}
\paragraph{Differentiable argmax.}
\label{sec:related_diffarg}
The argmax operator finds an index of the largest value in an array. CNNs involving the argmax operator cannot be trained via gradient-based optimizers, since it is not differentiable. Soft argmax~\cite{chapelle2010gradient} is a differentiable version of the argmax, and it has been applied to stereo matching~\cite{kendall2017end} and landmark detection~\cite{honari2018improving}. Incorporating the soft argmax operator with the Gumbel noise~\cite{gumbel1948statistical} relaxes a categorical distribution to a concrete one, enabling a differentiable sampling process~\cite{maddison2016concrete,jang2016categorical}. The soft argmax operator becomes the discrete one, as the temperature in the operator increases, but at the cost of unstable gradient flow. SFNet~\cite{lee2019sfnet} introduces a kernel soft argmax which combines a Gaussian kernel with the soft argmax, approximating the discrete argmax more accurately without using a large temperature parameter. We exploit the kernel soft argmax for network quantization. We propose to adjust the temperate parameter in the kernel soft argmax adaptively. As will be shown later, this addresses the quantizer gap problem, eliminating the discrepancies between the kernel soft argmax and the discrete one. 

\vspace{-0.2cm}
\section{Approach}
\vspace{-0.2cm}
In this section, we provide a brief description of our DAQ~(Sec.~\ref{sec:scheme}). We then describe each component of DAQ in detail, including DASR~(Sec.~\ref{sec:quant}) and a temperature controller~(Sec.~\ref{sec:temp}). 
\vspace{-0.2cm}
\subsection{Overview}
\vspace{-0.2cm}
\label{sec:scheme}
Using a uniform $b$-bit quantization, our quantizer~$Q$ maps full-precision inputs~$\hat{x}$, which could be weights~$\hat{w}$ or activations~$\hat{a}$, to quantized values~$Q(\hat{x})  \in \{0,1,..,2^b-1\}$ uniformly~(Fig.~\ref{fig:overview}). To this end, we first clip and normalize a full-precision input~$\hat{x}$ within a quantization interval, parameterized by upper~$u$ and lower~$l$ bounds~\cite{jung2019learning,gong2019differentiable}, as follows:
  \begin{equation}
      \label{eq:normalize}
    x = (2^b - 1) \frac{\mathrm{clip}(\hat{x},\min=l, \max=u) - l}{u - l},
  \end{equation}
where $x$ is a normalized input. We then apply DASR with a temperature controller, and rescale the output of DASR to assign the normalized input~$x$ to the nearest quantized value~$Q(\hat{x})$, which will be described in Sec.~\ref{sec:quant} and Sec.~\ref{sec:temp}, respectively. Finally, we scale quantized weights~$Q(\hat{w})$ or activations~$Q(\hat{a})$ linearly to the ranges of [-1,1] and [0,1], respectively, as follows:
\begin{equation}
    \label{eq:q_function}
    w_q=2\frac{Q(\hat{w})}{2^b-1} - 1, a_q =\frac{Q(\hat{a})}{2^b-1},
\end{equation}
where we denote by $w_q$ and $a_q$ elements of (scaled) quantized tensors for weights~${\bf{w}}_q$ and activations~${\bf{a}}_q$, respectively. With the quantized weights and activations, a convolutional output~${\bf{o}}$ is obtained as follows:
  \vspace{-0.2cm}
  \begin{equation}
  \label{eq:conv}
    {\bf{o}} = s{\bf{w}}_{q}*{\bf{a}}_{q},
  \vspace{-0.2cm}
  \end{equation}
  
where $*$ and $s$ are a convolutional operator and a learnable scalar parameter that adjusts the scale of the convolution output, respectively.

\subsection{DASR}
\vspace{-0.2cm}
\label{sec:quant}

The rounding function maps a full-precision input to its nearest quantized value. This assignment process can be thought of as the following two steps: First, distances between full-precision and quantized values are computed. Second, the nearest quantized value is chosen by applying the argmin operator over the distances, which is however not differentiable. Motivated by this, we propose DASR to approximate the discrete rounding function with a differentiable assignment operator. The approximation allows to use the same quantizer in both forward and backward passes, which alleviates the gradient mismatch problem. Similar to the two-step process, DASR takes a normalized input~$x$ in Eq.~\eqref{eq:normalize}, and computes distance scores w.r.t quantized values~$q \in \bf{q}$, where we denote by $\bf{q}$ a set of possible quantized values,~\ie,~$\{0,1,..,2^b-1\}$. It then assigns the input to a floating-point number, very close to the nearest quantized value, by a kernel soft argmax~\cite{lee2019sfnet}. In the following, we describe DASR in detail.

\vspace{-0.4cm}
\paragraph{Distance score.}
Given the normalized input~$x$, we compute distance scores for individual quantized values~$q \in \bf{q}$ as follows:
  \begin{equation}
  \label{eq:distance_score}
    d_{x}(q)=\exp(-|x-q|).
  \end{equation}
The distance score increases as the normalized input~$x$ becomes closer to the quantized value~$q$, and vice versa. Note that the computational cost of computing distance scores increases exponentially in accordance with the increase of the bit-widths,~\ie,~possible quantization values. To address this problem, we compute the distance scores only for the two nearest quantized values,~$q_f$ and~$q_c$, w.r.t the normalized input~$x$, where $q_f$ and~$q_c$ are obtained by $floor$ and $ceil$ functions, respectively, \ie.,~$q_c-q_f=1$.

\begin{figure}[t]
  \captionsetup{font={small}}
     \centering
         \includegraphics[width=\columnwidth]{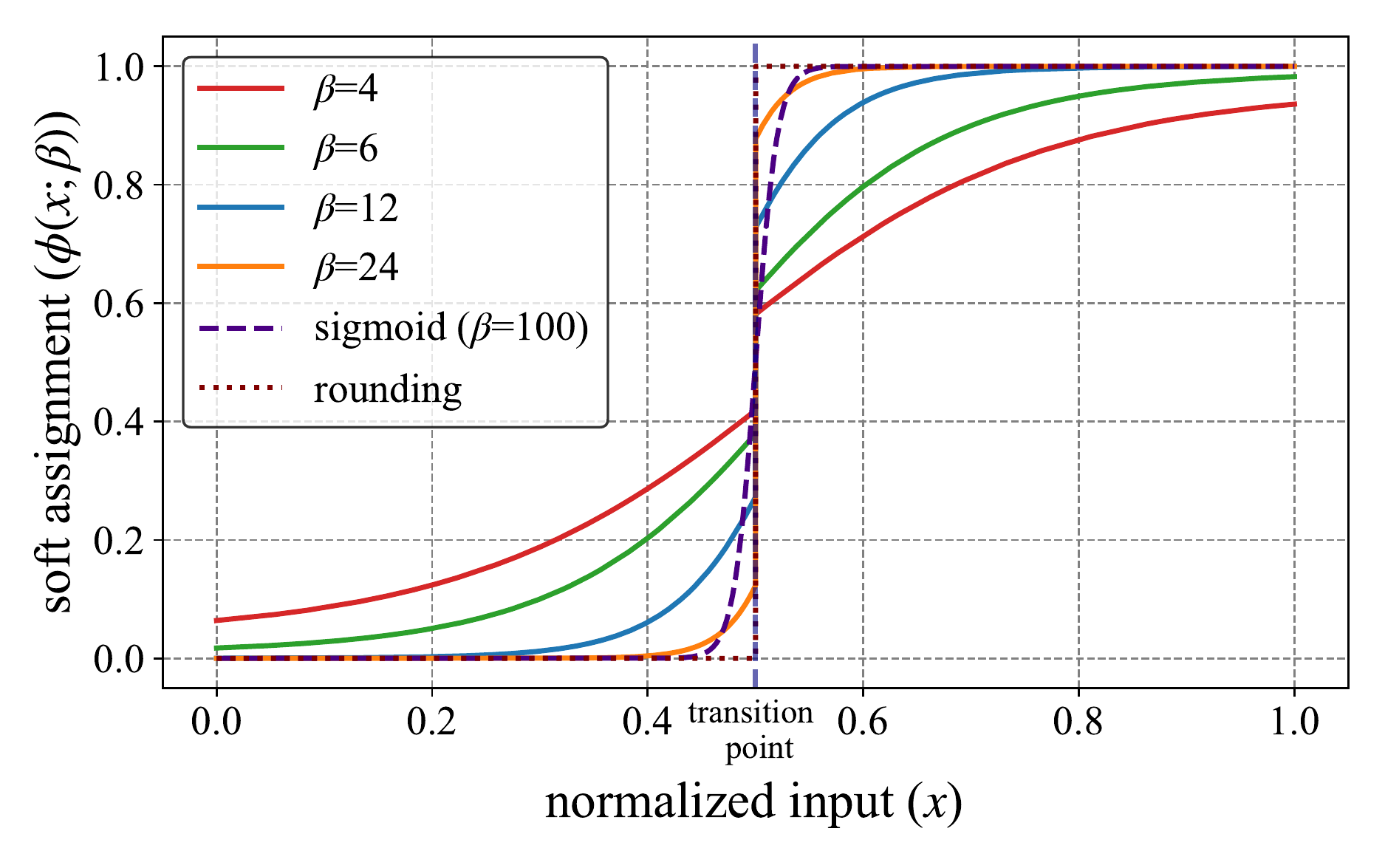} 
         \includegraphics[width=\columnwidth]{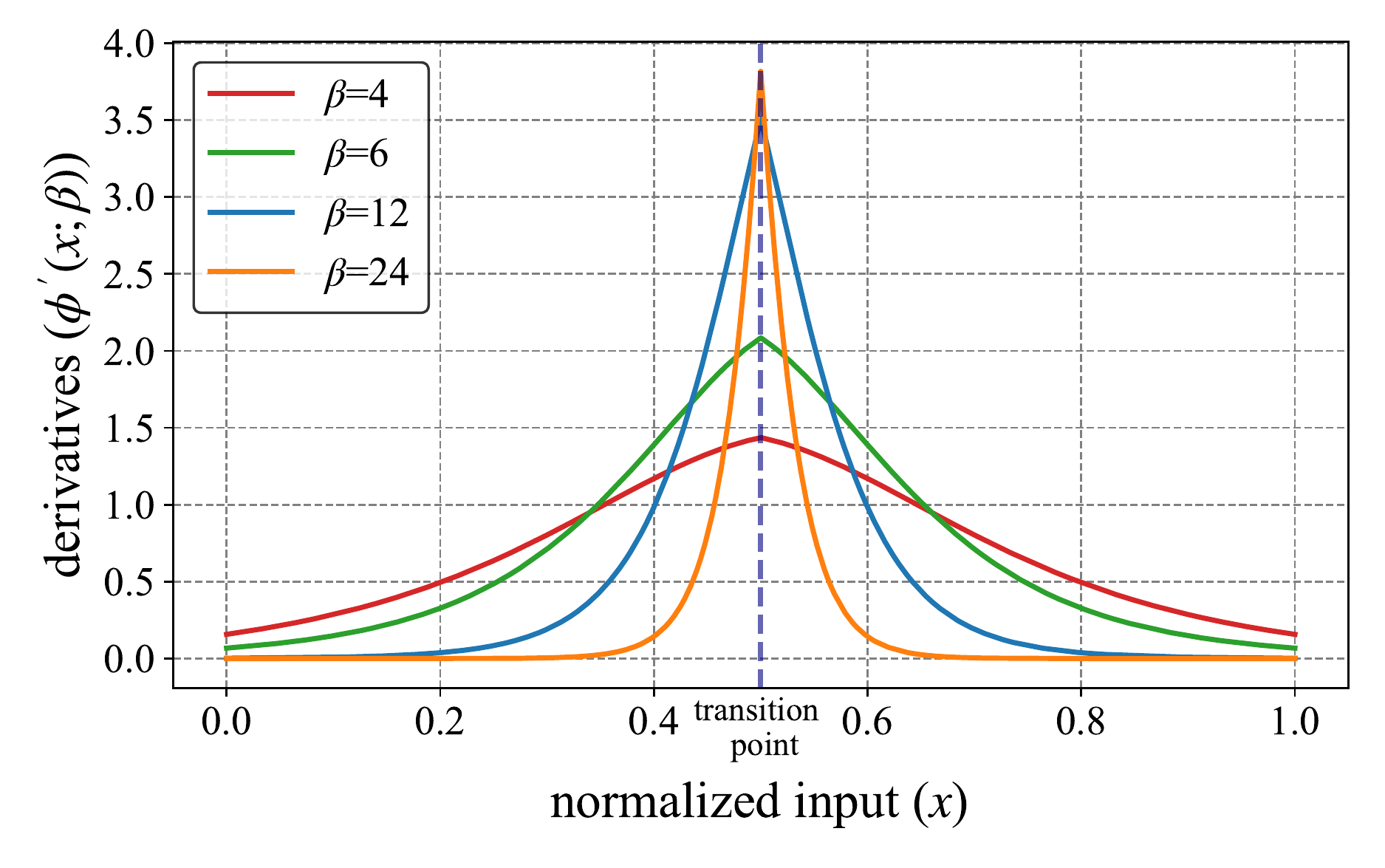}
       \vspace{-0.8cm}
       \label{fig:three sin x}
        \caption{Illustrations of soft assignments with different temperature parameters~$\beta$~(top) and corresponding derivatives~(bottom), for the case of the normalized input~$x$ being limited within the range of [0,1]. For comparison, we also plot a sigmoid function, exploited as a soft quantizer in~\cite{yang2019quantization}. We can see that the discrepancies between soft assignments and the rounding are significant especially near the transition point. They can be reduced by raising the temperature parameter. This, however, causes a vanishing gradient problem for the inputs near quantized values. (Best viewed in color.)}
        \label{fig:temp_relation}
  		\vspace{-0.4cm}
\end{figure}

\vspace{-0.4cm}
\paragraph{Soft assignment.} 
We can assign the normalized input~$x$ to the nearest quantized value by applying the discrete argmax function over the distance scores w.r.t the two quantized values,~$q_f$ and $q_c$, but this function is not differentiable. We instead propose to use the kernel soft argmax~\cite{lee2019sfnet} that approximates the discrete argmax, while maintaining differentiability. We define a soft assignment~$\phi(x;\beta)$ for the normalized input~$x$, with a temperature parameter~$\beta$, as an average of two quantized values,~$q_f$ and $q_c$, weighted by a distance probability~$m_{x}$~(Fig.~\ref{fig:temp_relation}(top)):
  \begin{equation}\label{eq:kernel_soft_max}
    \phi(x;\beta)=\sum_{i \in \{f,c\}}m_{x}(q_i;\beta) q_i.
  \end{equation}
The distance probability~$m_{x}$ is obtained by applying a softmax function, with the temperature parameter~$\beta$, to the distance scores~$d_{x}$ as follows:
  \begin{equation}
  \label{eq:distance_prop}
    m_{x}(q_i;\beta) = \frac{\exp(\beta k_{x}(q_i)d_{x}(q_i))}{\sum_{j \in \{f,c\} }\exp(\beta k_{x}(q_j)d_{x}(q_j))},
  \end{equation}
  \vspace{-0.2cm}
where we denote by $k_{x}$ a 1-dimensional Gaussian kernel centered on the nearest quantized value~(\ie,~$q_f$ or $q_c$) for the normalized input~$x$. The output of the kernel~$k_{x}(q)$ becomes larger as $q$ approaches to the nearest quantized value, which has an effect of retaining the distance score for the nearest quantized value, while suppressing the other one. This suggests that the distance probability is distributed with one clear peak around the nearest quantized value, and our soft assignment process approximates the discrete argmax well. For example, we can see from Fig.~\ref{fig:temp_relation}(top) that the soft assignment~$\phi$ approximates the discrete rounding more accurately than the sigmoid function adopted in soft quantization~\cite{yang2019quantization}, even with a much smaller temperature, preventing a gradient exploding problem. 

\vspace{-0.1cm}
\subsection{Temperature}
\vspace{-0.2cm}
\label{sec:temp}
The temperature parameter~$\beta$ adjusts a distribution of the distance probability~$m_x$, and it thus influences the soft assignment~$\phi$. The soft assignment with a fixed temperature parameter has the following limitations. (1) Small temperature parameters~(\eg,~$\beta=4$ in Fig.~\ref{fig:temp_relation}(top)) cause a quantizer gap problem, which is problematic particularly when a normalized input~$x$ is close to a transition point. (2) Large temperature parameters~(\eg,~$\beta=24$ in Fig.~\ref{fig:temp_relation}(top)) alleviate the quantizer gap problem. This, however, leads to a vanishing gradient problem. For example, the derivative of the soft assignment converges to zero rapidly, as the normalized input~$x$ moves away from a transition point in Fig.~\ref{fig:temp_relation}(bottom),~\ie,~approaches to a quantized value. Accordingly, exploiting a temperature parameter fixed for all inputs suffers from quantizer gap or vanishing gradient problems.
 
\begin{figure}[t]
  \captionsetup{font={small}}
     \centering
         \centering
         \includegraphics[width=\columnwidth]{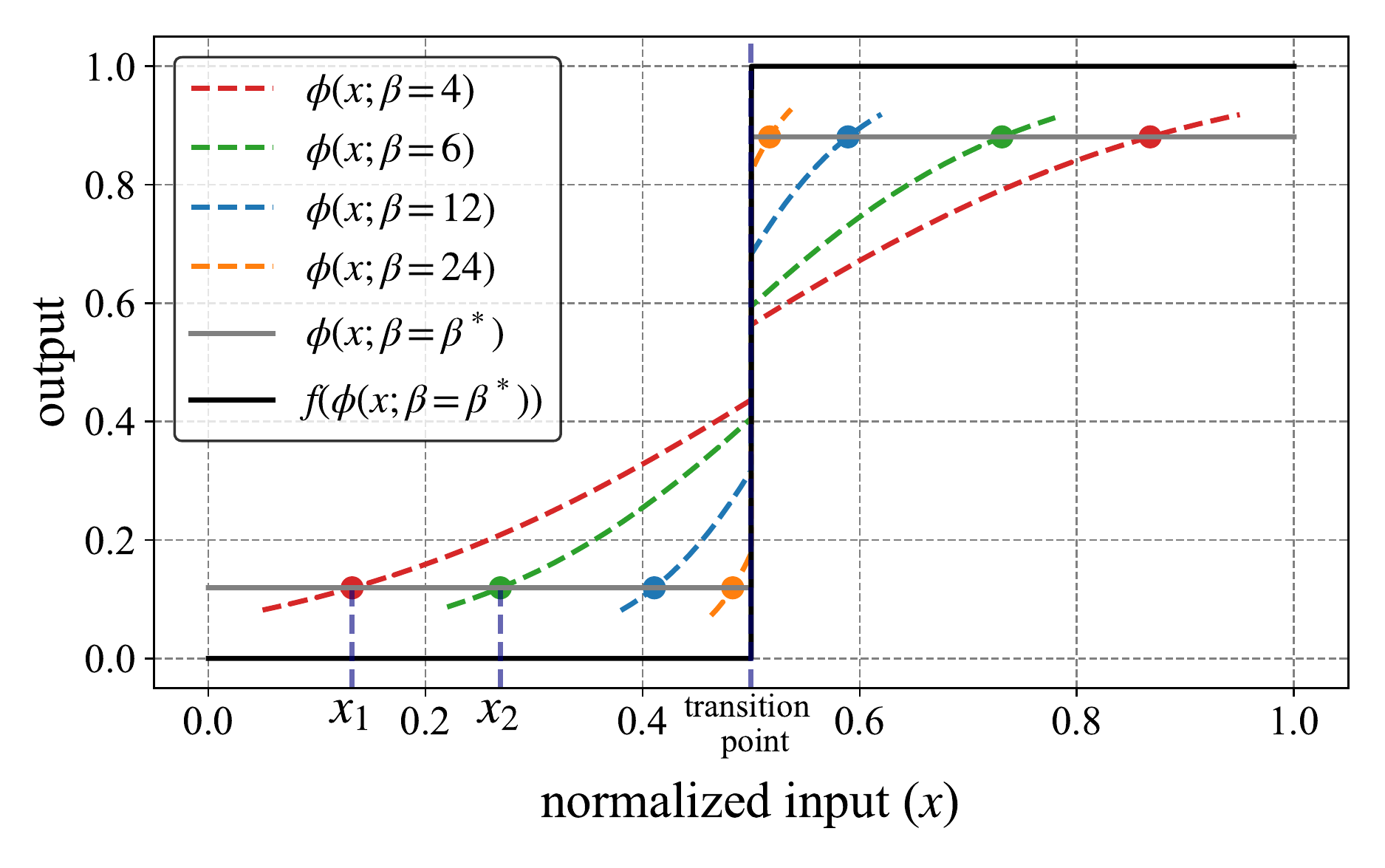}
       \vspace{-0.9cm}
        \caption{Illustrations of the output~$\phi(x;\beta=\beta^*)$ in Eq.~\eqref{eq:output_refined_temp} (gray line), and its rescaled version using a function~$f$ in~Eq.~\eqref{eq:rescale}~(black line). Computing the soft assignment with the adaptive temperature~$\beta^*$ has an effect of sampling output values~(colored circles) from different functions for individual inputs~(dotted lines). (Best viewed in color.)}
        \label{fig:rescale}
  \vspace{-0.6cm}
\end{figure}

\vspace{-0.4cm}
\paragraph{Temperature controller.} 
We adjust the temperature parameter~$\beta$ adaptively according to the distance between a normalized input and a transition point, such that we minimize the quantizer gap without suffering from the vanishing gradient problem. Specifically, we raise the temperature for the inputs near the transition point in order to address the quantizer gap problem. On the other hand, we lower the temperature for the inputs distant from the transition point, alleviating the vanishing gradient problem. To implement this idea, we define an adaptive temperature~$\beta^*$ as follows:
\begin{equation}
  \label{eq:control}
  \beta^*  = \frac{\gamma}{|s_{x}(q_f)-s_{x}(q_c)|},
\end{equation}
where $\gamma$ is a positive constant, and $s_{x}(q_i)$ is a weighted (distance) score defined as:
\vspace{-0.2cm}
\begin{equation}
	s_{x}(q_i) = k_{x}(q_i) d_{x}(q_i).
\end{equation}
As the input approaches to the transition point, the weighted scores, $s_{x}(q_f)$ and $s_{x}(q_c)$, become similar. In this case, the denominator in Eq.~\eqref{eq:control} decreases, and the adaptive temperature thus increases, alleviating the quantizer gap problem. On the contrary, the adaptive temperature decreases, when the input moves away from the transition point, avoiding the vanishing gradient problem. 

\begin{figure}[t]
  \captionsetup{font={small}}
     \centering
         \centering
         \includegraphics[width=\columnwidth]{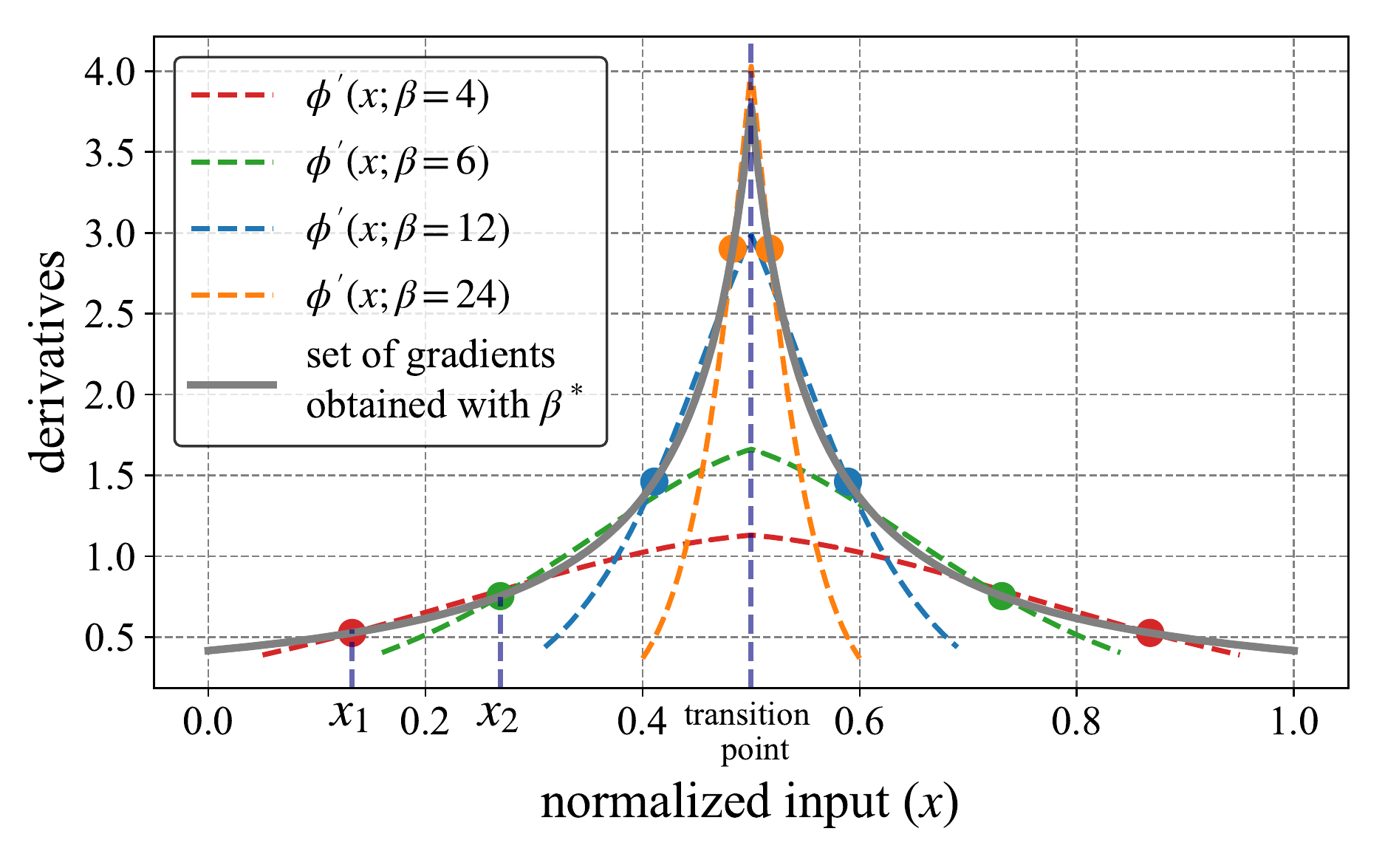}
       \vspace{-0.9cm}
        \caption{Illustrations of derivatives of soft assignment functions with different temperatures~(dotted lines), and a set of gradients for individual inputs~(gray line), obtained with different functions. The gradients for individual inputs are computed using the functions with corresponding adaptive temperature~$\beta^*$ (colored circles). (Best viewed in color.)}
        \label{fig:rescale_gradient}
  \vspace{-0.6cm}
\end{figure}


The adaptive temperature~$\beta^*$ varies according to the input, and the temperature in turn changes the shape of the soft assignment function (Fig.~\ref{fig:temp_relation}), suggesting that different soft assignment functions are applied for individual inputs. As an example, for the point of~$x_1$ in Fig.~\ref{fig:rescale}, where a value of the corresponding adaptive temperature is 4, its assignment is computed with the function of~$\phi(x;\beta=4)$. For the point of~$x_2$, the assignment is obtained with the function of~$\phi(x;\beta=6)$, which differs from the one used for~$x_1$ since the temperature is changed. By applying different functions for all inputs, we can obtain a set of assignments that can be represented analytically by plugging the adaptive temperature~$\beta^*$ into the soft assignment in Eq.~\eqref{eq:kernel_soft_max} as follows (see the supplement for details):

\begin{table*}[t]
  \setlength{\tabcolsep}{0.3em}
  \small
  \captionsetup{font={small}}
  \centering
  \begin{tabular}{l|C{0.7cm}|C{1.65cm}C{1.6cm}C{1.6cm}C{1.6cm}C{1.6cm}|C{1.6cm}C{1.6cm}C{1.6cm}}
  \multicolumn{1}{c|}{\multirow{2}{*}{Method}}& \multicolumn{9}{c}{Bit-width (W/A) } \\

   &FP & 1/1 & 1/2 & 2/2&  3/3 & \multicolumn{1}{c|}{4/4} & 1/32 & 2/32 & 3/32\\  

  \midrule   
  \midrule  
  LQ-Nets~\cite{zhang2018lq}         & 70.3  & - & 62.6 ($-$7.7) & 64.9 ($-$5.4) & 68.2 ($-$2.1) & 69.3 ($-$1.0) & - & 68.0 ($-$2.3) & 69.3 ($-$1.0) \\
  PACT~\cite{choi2018pact} & \bf{70.4} & - & 62.9 ($-$7.5) & 64.4 ($-$6.0) & 68.1 ($-$2.3) & 69.2 ($-$1.2) & - & 68.1 ($-$2.3) & 69.9 ($-$0.5) \\ 
  QIL~\cite{jung2019learning} & 70.2 & - & - & 65.7 ($-$4.5) & 69.2 ($-$1.0) & 70.1 ($-$0.1) & 65.8 ($-$4.4) & - & - \\ 
  QNet \cite{yang2019quantization} & 70.3 & 53.6 ($-$16.7) & 63.4 ($-$6.9) & - & - & \multicolumn{1}{c|}{-} & 66.5 ($-$3.8) & 69.1 ($-$1.2) & 70.4 ($+$0.1) \\ 
  RQ~\cite{louizos2018relaxed} & 69.5 & - & - & - & - & 62.5$^\dagger$\hspace{-0.05cm}($-$7.0) & - & - & - \\ 
  DSQ~\cite{gong2019differentiable} & 69.9 & - & - & 65.2 ($-$4.7) & 68.7 ($-$1.2) & 69.6$^\dagger$\hspace{-0.05cm}($-$0.3) & - & - & - \\ 
  LSQ~\cite{esser2019learned} & 70.1 & - & - & 66.8 ($-$3.3) & 69.3 ($-$0.8) & 70.7 ($+$0.6) & - & - & - \\ 
  LSQ+~\cite{bhalgat2020lsq+} & 70.1 & - & - & 66.7 ($-$3.4) & 69.4 ($-$0.7) & \bf{70.8 ($+$0.7)} & - & - & - \\ 
  IRNet \cite{qin2020forward} & 69.6 & - & - & - & - & \multicolumn{1}{c|}{-} & 66.5 ($-$3.1) & - & - \\ 
  Ours          & 69.9  & \bf{56.2} ($-$13.7) & \bf{64.6} ($-$5.3) & \bf{66.9} ($-$3.0) & \bf{69.6} ($-$0.3) & 70.5 ($+$0.6) & \bf{67.2} ($-$2.7)     & \bf{69.8} ($-$0.1) & \bf{70.8} ($+$0.9) \\ 
  \end{tabular}
  \vspace{-0.3cm}
  \caption{Quantitative results of ResNet-18~\cite{he2016deep} on the validation split of ImageNet~\cite{deng2009imagenet}. We report the top-1 accuracy for comparison. We denote by ``W'' and ``A'' the bit-precision of weights and activations, respectively. ``FP'' and $\dagger$ represent accuracies for full-precision and fully quantized models, respectively. Numbers in bold indicate the best performance. Numbers in parentheses are accuracy improvements or degradations compared to the full-precision one.}
  
  \label{tab:resnet18}
  \vspace{-0.6cm}
\end{table*}

\vspace{-0.2cm}
\begin{equation}
  \label{eq:output_refined_temp}
  \phi(x;\beta=\beta^*) = \begin{cases} q_f + \lambda , &x \leq q_t \\
                                q_c - \lambda, & x > q_t, \end{cases}
\end{equation}
where $\lambda= 1 / {(e^\gamma+1)}$, and we denote by $q_t$ a transition point, defined as $(q_f+q_c) / {2}$. We can see from Fig.~\ref{fig:rescale} that the set of assignments, sampled from different soft assignment functions for individual inputs, are the same as output values of a single discrete function (Fig.~\ref{fig:rescale}(gray line)). Analogous to the process of obtaining the set of assignments in a forward pass, the gradients for individual inputs are obtained with different functions for backward propagation~(Fig.~\ref{fig:rescale_gradient}). That is, the gradient for each input is obtained from the soft assignment function with the corresponding adaptive temperature. For example, the gradients for the points,~$x_1$ and~$x_2$ in Fig.~\ref{fig:rescale_gradient}, are computed using the derivatives of the corresponding functions,~$\phi'(x;\beta=4)$ and $\phi'(x;\beta=6)$, respectively. This is effective in alleviating the vanishing gradient problem, which is particularly severe, when the temperature is set to a large value~(\eg,~$\beta=24$ in Fig.~\ref{fig:temp_relation}(top)). Note that the adaptive temperature~$\beta^*$ is regarded as a hyperparameter for each input in both forward and backward passes. In summary, leveraging a different function for each input enables computing the gradient for backward propagation, while providing the outputs that coincide with those obtained from a discrete rounding but with offsets of~$\lambda$~(Fig.~\ref{fig:rescale}). Rescaling the output in Eq.~\eqref{eq:output_refined_temp} with the following function,
\vspace{-0.15cm}
\begin{equation}
  \label{eq:rescale}
  f(y) = \frac{y-q_t}{1-2\lambda}+q_t,
\vspace{-0.15cm}
\end{equation}
we can obtain 
\vspace{-0.15cm}
\begin{equation}
  \label{eq:rescaled_output}
  f(\phi(x;\beta=\beta^*)) = \begin{cases} q_f, & x \leq q_t \\
    q_c, &x > q_t, \end{cases}
\vspace{-0.15cm}
\end{equation}
which corresponds to the output of DAQ~(\ie.,~$Q(\hat{x})$). It is clear that the rescaled output in Eq.~\eqref{eq:rescaled_output} provides the exactly same values as the rounding function, suggesting that our DAQ is free from the quantizer gap problem, even using the rounding at test time (see the last row in Table.~\ref{tab:temp}). We summarize the overall quantization process in the supplement.
\vspace{-0.7cm}
\section{Experiments}
\vspace{-0.2cm}

\subsection{Experimental details}
\vspace{-0.15cm}
\label{sec:detail}
\paragraph{Implementation details.}
We quantize weights and/or activations for ResNets~\cite{he2016deep}~(\ie, ResNet-18, -20, and -34) and MobileNet-V2~\cite{sandler2018mobilenetv2}. Following~\cite{jung2019learning,park2020profit}, we do not quantize the first and last layers for all network architectures except for MobileNet-V2, where all layers are quantized. We empirically set the constant~$\gamma$ in Eq.~\eqref{eq:control} to 2 for both weight and activation quantizers, and the standard deviation of the Gaussian kernel~$k_x$ to 1 and 2 for quantizers of weight and activation, respectively. We use a grid search to set these parameters. We choose the ones that give the best performance on the validation split\footnote{We divide the training split of CIFAR-10~\cite{krizhevsky2009learning} into training and validation sets for the grid search.} of CIFAR-10~\cite{krizhevsky2009learning}, and fix them for all experiments.

\vspace{-0.4cm}
\paragraph{Training.}
Network weights are trained using the SGD optimizer with learning rates of 1e-2 and 5e-3 for ResNets and MobileNet-V2, respectively. We learn quantization parameters, such as the lower and upper bounds, $l$ and $u$ in Eq.~\eqref{eq:normalize}, and the scale factor~$s$ in Eq.~\eqref{eq:conv}, using the Adam optimizer~\cite{kingma2014adam} with a learning rate of 1e-4. The learning rates for all parameters are scheduled by the cosine annealing strategy~\cite{loshchilov2016sgdr}. For the ResNet-20 architecture, we train the quantized networks for 400 epochs on CIFAR-10~\cite{krizhevsky2009learning} with a batch size of 256, and the weight decay is set to 1e-4. Other networks are trained for 100 epochs on ImageNet~\cite{deng2009imagenet} with batch sizes of 256 and 160 for ResNets~(\ie., ResNet-18 and -34) and MobileNet-V2, respectively. For ResNet-18 and -34, the weight decay is set to 1e-4, except for low-bit quantizations (\ie.,~1/1, 1/2, and 2/2-bit settings), where we use a smaller weight decay of 5e-5, following~\cite{esser2019learned}. For MobileNet-V2, the weight decay is set to 4e-5. We do not use weight decay for learning quantization parameters.
\vspace{-0.5cm}
\paragraph{Initialization.}
The weights in all quantized networks are initialized from the full-precision pretrained models. We apply standardization to the weights~\cite{li2019additive} before feeding them into quantizers. We initialize lower and upper bounds in the weight quantizer to -3 and 3, respectively. Lower and upper bounds in the activation quantizer are initialized by~$-3\sigma_{A}$ and~$3\sigma_{A}$, respectively, where~$\sigma_{A}$ is a standard deviation of input activations in a layer, except when an input of the quantizer is pre-activated by a ReLU. In this case, we fix a lower bound to zero, and learn an upper bound only with an initialization of~$3\sigma_{A}$.

\begin{table*}[t] 
  \setlength{\tabcolsep}{0.3em}
  \small
  \captionsetup{font={small}}
  \centering
  \begin{tabular}{l|C{1cm}|C{1.83cm}C{1.83cm}C{1.83cm}C{1.83cm}C{1.83cm}|C{1.83cm}}
  \multicolumn{1}{c|}{\multirow{2}{*}{Method}}& \multicolumn{6}{c}{Bit-width (W/A) } \\

   &FP & 1/1 & 1/2 & 2/2 & 3/3&  4/4 & 1/32\\  

  \midrule   
  \midrule  
  LQ-Nets \cite{zhang2018lq} & \bf{73.8}  & - & 66.6 ($-$7.2) & 69.8 ($-$4.0) & 71.9 ($-$1.9) & - & - \\ 
  QIL~\cite{jung2019learning} & 73.7 & - & - & 70.6 ($-$3.1) & {\bf{73.1}} ($-$0.6) & {~\bf{73.7}} ($+$0.0) & -\\
  DSQ~\cite{gong2019differentiable} & 73.3 & - & - & 70.0 ($-$3.3) & 72.5 ($-$0.8) & ~72.8 ($-$0.5) & - \\
  IRNet~\cite{qin2020forward} & 73.3 & - & - & - & - & - & 70.4 ($-$2.9) \\
  Ours          & 73.3 & \bf{62.1 ($-$11.2)} & \bf{69.4 ($-$3.9)} & \bf{71.0} ($-$2.3) & \bf{73.1} ($-$0.2) & ~\bf{73.7 ($+$0.4)} & \bf{71.9 ($-$1.4)}\\ 

  \end{tabular}
  \vspace{-0.3cm}
  \caption{Quantitative results of ResNet-34~\cite{he2016deep} on the validation split of ImageNet~\cite{deng2009imagenet}. We report the top-1 accuracy for comparison. W/A: Bit-precision of weights/activations; FP: Results obtained by full-precision models.}
  \label{tab:resnet34}
  \vspace{-0.6cm}
\end{table*}

\begin{table}[t] 
  \setlength{\tabcolsep}{0.3em}
  \small
  \captionsetup{font={small}}
  \centering
  \begin{tabular}{l|C{0.9cm}|C{1.95cm}}
  \multicolumn{1}{c|}{\multirow{2}{*}{Method}}& \multicolumn{2}{c}{Bit-width (W/A) } \\

   &FP & \multicolumn{1}{c}{4/4}\\  

  \midrule   
  \midrule  
  PACT \cite{choi2018pact} & 71.8  & ~61.4 ($-$10.4) \\ 
  DSQ~\cite{gong2019differentiable} & \bf{71.9}  & ~64.8 ~~($-$7.1) \\ 
  PROFIT \cite{park2020profit} & \bf{71.9}  & ~\bf{71.6}\hspace{0.03cm}$^\dagger$ ($-$0.3) \\ 
  Ours          & \bf{71.9}  & ~70.0\hspace{0.03cm}$^\dagger$ ($-$1.9)\\ 

  \end{tabular}
  \vspace{-0.3cm}
  \caption{Quantitative results of MobileNet-V2~\cite{sandler2018mobilenetv2} on the validation split of ImageNet~\cite{deng2009imagenet}. We report the top-1 accuracy for comparison. W/A: Bit-precision of weights/activations; FP: Results obtained by full-precision models; $\dagger$: Results from fully quantized models.}
  \vspace{-0.6cm}
  \label{tab:mobile}
\end{table}

\vspace{-0.2cm}
\subsection{Results}
\vspace{-0.2cm}
We evaluate our approach with network architectures, including ResNet-18, -20, -34~\cite{he2016deep}, and MobileNet-V2~\cite{sandler2018mobilenetv2}, for various bit-widths, and compare the performance with the state of the art for image classification  on CIFAR-10~\cite{krizhevsky2009learning} and ImageNet~\cite{deng2009imagenet}.

\vspace{-0.5cm}
\paragraph{ImageNet.}
The ImageNet dataset~\cite{deng2009imagenet} provides approximately 1.2 million training and 50K validation images of 1,000 categories with corresponding ground-truth class annotations. We train and evaluate a quantized network on the training and validation splits, respectively. We use the top-1 accuracy to quantify the performance.

We show in Table~\ref{tab:resnet18} the top-1 accuracy on ResNet-18, and compare our approach with the state of the art. All numbers in Table~\ref{tab:resnet18} are taken from each paper, except for LSQ~\cite{esser2019learned}\footnote{It uses a pre-activated version of ResNet, which is different from the standard architecture. We take the numbers from the work of LSQ+~\cite{bhalgat2020lsq+}, where LSQ is reproduced with the standard ResNet.}. We observe five things from this table: (1) Our method outperforms the state of the art by a significant margin in terms of the top-1 accuracy especially for low-bit quantizations~(\ie.,~1/1, 1/2, 1/32, 2/32-bit settings). LSQ~\cite{esser2019learned} and LSQ+~\cite{bhalgat2020lsq+} show better results than ours slightly in a 4/4-bit setting, but using a more accurate full-precision model. The high-performance model provides a better initialization to optimize quantized networks. (2) Our approach is effective to binarize networks~(\ie.,~1/1 and 1/32-bit  settings), outperforming IRNet~\cite{qin2020forward}\footnote{It shows top-1 accuracy of 58.1 in an 1/1-bit setting, but using ResNet-18 with the Bi-Real structure~\cite{liu2018bi} that adds additional residual connections. For fair comparison, we only report the results of IRNet using the same network architectures as ours} designed for network binarization. This also demonstrates the effectiveness of DASR on alleviating the gradient mismatch problem. As stated in~\cite{lin2016overcoming}, the gradient mismatch problem becomes even worse, as the bit-width of weights and/or activations is small. (3) Ours performs better than soft quantizers~\cite{yang2019quantization,gong2019differentiable,louizos2018relaxed}, even they also alleviate the gradient mismatch problem. This indicates that addressing the quantizer gap problem improves the performance significantly. (4) Our method gives better results than other quantizers, similar to ours, that learn either quantization intervals~\cite{jung2019learning,gong2019differentiable,esser2019learned} or clipping ranges of activations~\cite{choi2018pact} using the STE. (5) We can employ our quantization method in various bit-widths, and achieve the state-of-the-art performance consistently, while others may apply for specific settings.

Tables~\ref{tab:resnet34} and~\ref{tab:mobile} show quantitative comparisons with the state of the art using ResNet-34~\cite{he2016deep} and MobileNet-V2~\cite{sandler2018mobilenetv2}, respectively. Our method outperforms the state of the art in low-bit quantizations (\eg.,~1/1, 2/2, and 1/32-bit), and shows the same accuracy as QIL~\cite{jung2019learning} in 3/3 and 4/4-bit settings. QIL uses a progressive learning technique~\cite{zhuang2018towards} training a quantized network sequentially from high- to low-bit precision, which is computationally demanding. In contrast, our approach fine-tunes a full-precision model directly to achieve quantized networks with the target precision. Note that our method shows the best performance in terms of an accuracy improvement/degradation from a full-precision model. Table~\ref{tab:mobile} demonstrates that our method is also effective to quantize a light-weight network architecture (\ie.,~MobileNet-V2). Note that PROFIT~\cite{park2020profit} is specially designed for quantizing the light-weight network architectures, and also uses many heuristics (\eg.,~progressive learning~\cite{zhuang2018towards,jung2019learning} and distillation~\cite{zhuang2018towards,polino2018model}) at training time, which requires more training time and computational cost, compared to our approach.

\begin{table}[t] 
  \setlength{\tabcolsep}{0.3em}
  \small
  \captionsetup{font={small}}
  \centering
  \begin{tabular}{l|C{0.9cm}|C{1.9cm}|C{1.9cm}}
  \multicolumn{1}{c|}{\multirow{2}{*}{Method}}& \multicolumn{3}{c}{Bit-width (W/A)} \\

   &FP & \multicolumn{1}{c|}{1/1} & 1/32\\  

  \midrule   
  \midrule  
  DoReFa \cite{zhou2016dorefa}        & 90.8      & ~79.3 ($-$11.5)    & 90.0 ($-$0.8) \\ 
  LQ-Net~\cite{zhang2018lq}            & \bf{92.1}      & \multicolumn{1}{c|}{-}                 & 90.1 ($-$2.0) \\ 
  DSQ~\cite{gong2019differentiable}    & 90.8      & ~84.1 ~~($-$6.7)     & 90.2 ($-$0.6) \\ 
  IRNet~\cite{qin2020forward}        & 91.7    & ~85.4 ~~($-$6.3)    & 90.8 ($-$0.9) \\ 
  Ours                                & 91.4     & ~\bf{85.8 ~~($-$5.6)}    & \bf{91.2 ($-$0.2)} \\ 

  \end{tabular}
  \vspace{-0.3cm}
  \caption{Quantitative results of ResNet-20~\cite{he2016deep} on the test split of CIFAR-10~\cite{krizhevsky2009learning}. We report the top-1 accuracy for comparison. W/A: Bit-precision of weights/activations; FP: Results obtained by full-precision models.}
  \vspace{-0.5cm}
  \label{tab:resnet20}
\end{table}
\vspace{-0.4cm}
\paragraph{CIFAR-10.}
The CIFAR-10 dataset~\cite{krizhevsky2009learning} consists of 50k training and 10K test images of 10 object categories for image classification. We train a quantized network with the training set, and report the top-1 accuracy on the test split. We show in Table~\ref{tab:resnet20} a quantitative comparison with the state of the art using the ResNet-20 architecture~\cite{he2016deep}. We can clearly see that our method outperforms the state of the art, confirming the effectiveness of our approach again.

\label{sec:performance}

\vspace{-0.2cm}
\subsection{Discussion}
\vspace{-0.2cm}
\label{sec:ablation}
We present an ablation analysis on DASR and compare our temperature controller with other methods alleviating the quantizer gap problem. We report the top-1 accuracy with 1/1-bit ResNet-20~\cite{he2016deep} on the test split of CIFAR-10~\cite{krizhevsky2009learning}. More analysis on DAQ can be found in the supplement.

\begin{table}
\setlength{\tabcolsep}{0.3em}
  \small
\captionsetup{font={small}}
\begin{center}
\begin{tabular}{c|C{2.1cm}| C{1.9cm} | C{1.1cm}}
\multirow{2}{*}{Type} & Temperature & \multicolumn{2}{c}{Test time}    \\
& ($\beta$) & Rounding   & DASR    \\
\cmidrule{1-4}\morecmidrules\hline
\multirow{4}{*}{\rb{Soft~}} \multirow{4}{*}{\rb{argmax~}}
&& \\[-0.9em]
& 10        &    11.7 ($-$75.1) &    86.8         \\
& 20        &    10.2 ($-$44.4) &    54.6         \\
& 60        &    10.0 ($-$27.3) &    37.3         \\
& 150        &    -        &    - \\
\hline
\multirow{5}{*}{\rb{Kernel soft~}} \multirow{5}{*}{\rb{argmax~}}
&& \\[-0.9em]
& 4        &    13.5 ($-$76.2)  &    \bf{89.7}         \\
& 8        &    48.8 ($-$37.2)  &    86.0         \\
& 12        &    69.9 ($-$10.9) &    80.8         \\
& 24        &    59.6 ~~($-$3.8) &    63.8         \\
& $\beta^*$ &    \bf{85.8} ~~($-$0.0)  & 85.8        \\
\hline
\end{tabular}
\captionsetup{font={small}}
\vspace{-3mm}
\caption{Quantitative comparison for variants of our method with different argmax operators and temperature parameters. We also report the results obtained by DASR at test time, instead of the rounding function, that is, we use the same quantization function for both training and test time to quantify the influence of the quantizer gap problem on quantization. Note that weights and activations in this case, except the result for the adaptive temperature~$\beta^*$, are full-precision values, not quantized ones. Numbers in parentheses are accuracy drops between full-precision and quantized models.}
\vspace{-0.7cm}
\label{tab:temp}
\end{center}
\end{table}

\begin{figure}[t]
  \captionsetup{font={small}}
     \centering
     \includegraphics[width=0.85\columnwidth]{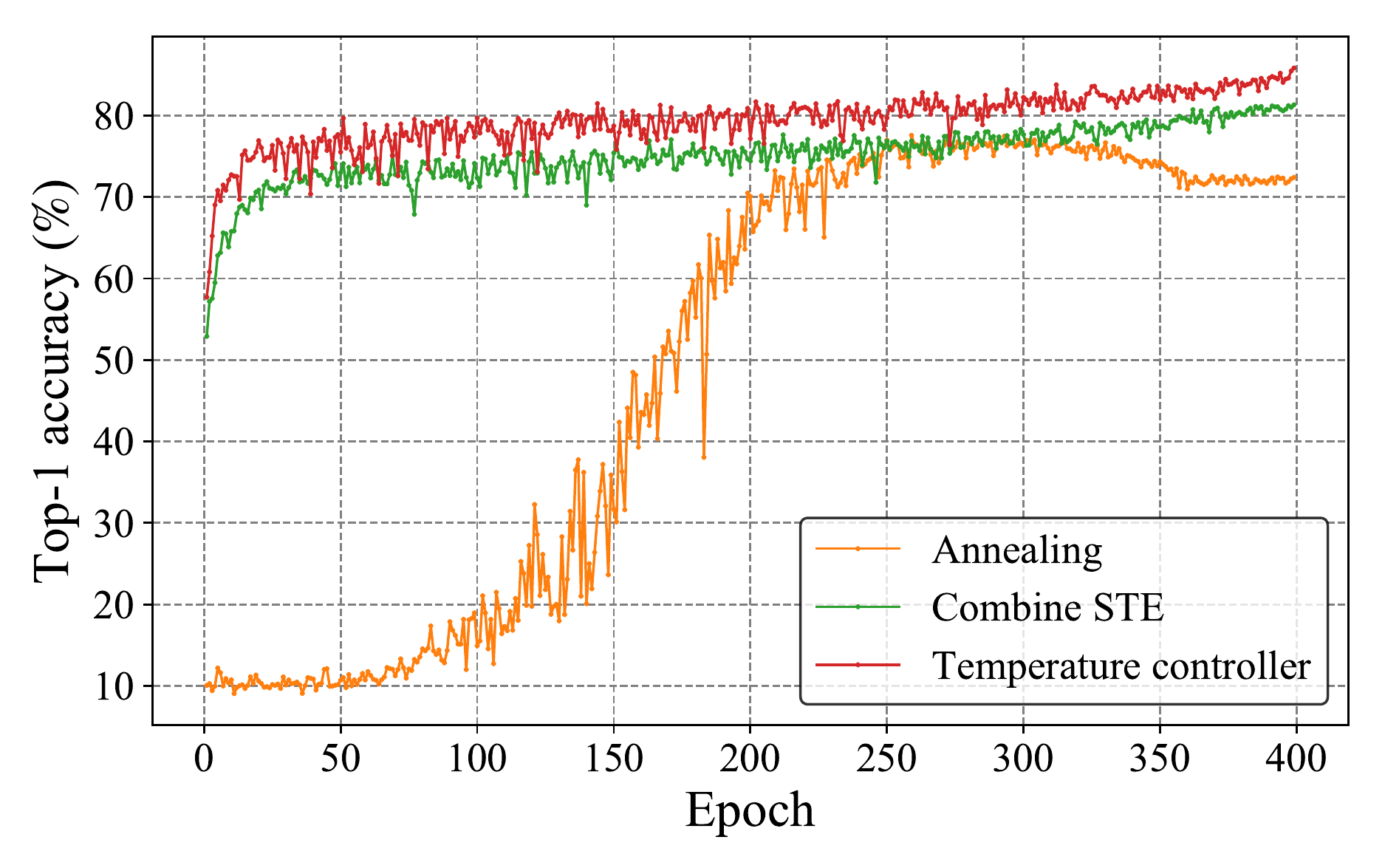}
       \vspace{-0.4cm}
     \caption{Visual comparison of top-1 accuracies for different methods alleviating the quantizer gap problem. (Best viewed in color.)}
  \label{fig:accuracy}
  \vspace{-0.5cm}  
\end{figure}

\paragraph{Differentiable argmax.}
We compare in the third column of Table~\ref{tab:temp} the quantization performance for variants of our method. We use DASR to train quantized networks with different argmax operators and temperature parameters, and exploit the rounding function as a discretizer at test time. To quantify the influence of the quantizer gap problem, we also report the results when exploiting DASR at test time in the fourth column of Table~\ref{tab:temp}, such that we use the same quantization function for both training and test time. Note that DASR with a fixed temperature outputs floating-point numbers as described in Sec.~\ref{sec:temp}, indicating that the results in the fourth column, except for the one for the adaptive temperature~$\beta^*$, are obtained with full-precision weights and activations, not quantized ones. We can see from the first four rows that an accuracy drop caused by the quantizer gap decreases, according to the increase of the temperature parameter. This, however, results in an unstable gradient flow, degrading the performance, even for the cases without the quantizer gap~(\eg,~86.8 for $\beta$=10 vs. 37.3 for $\beta$=60). We fail to train the network with $\beta=150$, due to a gradient exploding problem. The last five rows show that the kernel soft argmax~\cite{lee2019sfnet} is more effective to handle the quantizer gap problem than the soft argmax, even with a much smaller temperature parameter. For example, raising the temperature parameter from 4 to 24 reduces the accuracy drop from 76.2\% to 3.8\%. Exploiting a large temperature ($\beta=24$), however, causes a vanishing gradient problem as stated in Sec.~\ref{sec:temp}, which degrades the quantization performance. From the last row, we can observe that the adaptive temperature addresses the quantizer gap and vanishing gradient problems, providing the best result without the performance drop.
\vspace{-0.5cm}
\paragraph{Temperature controller.}
Table~\ref{tab:perform_mis} compares our temperature controller with other methods~\cite{yang2019quantization,gong2019differentiable,louizos2018relaxed} for avoiding the quantizer gap problem. For fair comparison, we adopt these methods within our DASR framework. Specifically, for the temperature annealing~\cite{yang2019quantization}, we raise the temperature parameter from 2 to 48 gradually at training time, such that DASR approaches to the rounding function. To combine the STE with DASR~\cite{gong2019differentiable,louizos2018relaxed}, we use the discrete rounding for a forward pass, while using the derivative of DASR in a backward pass. Exploiting the temperature controller takes more time than other methods due to the additional computations of adjusting the temperature~$\beta$ for individual inputs~$x$. This, however, alleviates the quantizer gap and gradient mismatch problems jointly, outperforming other methods by a large margin. We compare in Fig.~\ref{fig:accuracy} top-1 accuracy curves. We can observe that our temperature controller gives better results compared with others during training. 
\changed{}

\begin{table}
\setlength{\tabcolsep}{0.3em}
  \small
\captionsetup{font={small}}
\begin{center}

\begin{tabular}{L{3.7cm} |C{1.4cm} |C{2.1cm}}
                
\multicolumn{1}{c|}{Method}   & Time/iters (\emph{ms}) & Top-1 accuracy (\%)    \\

\cmidrule{1-3}\morecmidrules\hline
& \\[-0.9em]
Annealing~\cite{yang2019quantization} & 286.2 &    72.2 \\
Combine STE~\cite{gong2019differentiable,louizos2018relaxed}  &	287.8 &    81.7     \\
Temperature Controller  & 296.9	&    \bf{85.8} \\
\hline
\end{tabular}
\captionsetup{font={small}}
\vspace{-3mm}
\caption{Training time and accuracy comparisons of our temperature controller with other methods avoiding the quantizer gap problem. The time is measured with an RTX 2080Ti GPU.}
\vspace{-0.7cm}
\label{tab:perform_mis}
\end{center}
\end{table}

\vspace{-0.2cm}
\section{Conclusion}
\vspace{-0.2cm}
We have shown that network quantization can be formulated as an assignment problem between full-precision and quantized values, and introduced a novel quantizer, dubbed DAQ, that addresses both the gradient mismatch and quantizer gap problems in a unified framework. Specifically, DASR approximates a rounding function with a kernel soft argmax operator, together with a temperature controller adjusting the temperature parameter adaptively. We have shown that DAQ achieves the state of the art for various network architectures and bit-widths without bells and whistles. We have also verified the effectiveness of each component of DAQ with a detailed analysis.

\noindent \textbf{Acknowledgments.}
This research was supported by the Samsung Research Funding \& Incubation Center for Future Technology (SRFC-IT1802-06).

{\small
\bibliographystyle{ieee_fullname}
\bibliography{egbib}
}

\clearpage


\section*{Supplementary material}

# Distance-aware Quantization
# Supplement

Dohyung Kim    Junghyup Lee    Bumsub Ham*
School of Electrical and Electronic Engineering, Yonsei University
https://cvlab.yonsei.ac.kr/projects/DAQ

In this supplementary material, we present a detailed derivation of an output in Eq. (9) of our main paper, and an overall quantization process of DAQ in Sec. 1 and 2, respectively. We also provide more analysis on DAQ including the design and experimental results on the super-resolution (SR) task in Sec. 3 and 4, respectively.

## 1. Derivation of output in Eq. (9)

The output in Eq. (9) is obtained by plugging the adaptive temperature $\beta^*$ into the soft assignment function $\phi$ in Eq. (5) of the main paper. The soft assignment is defined as a weighted average using two nearest quantized values, $q_f$ and $q_c$, and corresponding distance probabilities $m_x$ as follows (See Eqs. (5-6) in the main paper):

$$\phi(x; \beta = \beta^*) = m_x(q_f; \beta = \beta^*)q_f + m_x(q_c; \beta = \beta^*)q_c$$
$$= \frac{q_f \exp(\beta^* s_x(q_f)) + q_c \exp(\beta^* s_x(q_c))}{\exp(\beta^* s_x(q_f)) + \exp(\beta^* s_x(q_c))}$$
$$= \frac{q_f \exp(\beta^*(s_x(q_f) - s_x(q_c))) + q_c}{\exp(\beta^*(s_x(q_f) - s_x(q_c))) + 1}. \quad (1)$$

We define the adaptive temperature $\beta^*$ as follows:

$$\beta^* = \frac{\gamma}{|s_x(q_f) - s_x(q_c)|}, \quad (2)$$

which can be expressed as:

$$\beta^* = \begin{cases} \dfrac{\gamma}{s_x(q_f) - s_x(q_c)}, & x \leq q_t \\ \dfrac{-\gamma}{s_x(q_f) - s_x(q_c)}, & x > q_t, \end{cases} \quad (3)$$

since the weighted score of $q_f$ is larger than that of $q_c$, i.e., $s_x(q_f) > s_x(q_c)$, when the normalized input $x$ is less than the transition point $q_t$, defined as $(q_f + q_c)/2$, and vice versa. Plugging the adaptive temperature $\beta^*$ in Eq. (3) into the soft assignment $\phi$ in Eq. (1), we obtain the following results:

$$\phi(x; \beta = \beta^*) = \begin{cases} \dfrac{q_f \exp(\gamma) + q_c}{\exp(\gamma) + 1}, & x \leq q_t \\ \dfrac{q_f \exp(-\gamma) + q_c}{\exp(-\gamma) + 1}, & x > q_t \end{cases}$$
$$= \begin{cases} \dfrac{q_f \exp(\gamma) + q_c}{\exp(\gamma) + 1}, & x \leq q_t \\ \dfrac{q_c \exp(\gamma) + q_f}{\exp(\gamma) + 1}, & x > q_t \end{cases} \quad (4)$$
$$= \begin{cases} (q_f \exp(\gamma) + q_c)\lambda, & x \leq q_t \\ (q_c \exp(\gamma) + q_f)\lambda, & x > q_t, \end{cases}$$

where $\lambda = 1/(\exp(\gamma) + 1)$ as stated in Sec. 3.3 of the main paper. We then reformulate this equation as follows:

$$\phi(x; \beta = \beta^*) = \begin{cases} (q_f \exp(\gamma) + q_f - q_f + q_c)\lambda, & x \leq q_t \\ (q_c \exp(\gamma) + q_c - q_c + q_f)\lambda, & x > q_t \end{cases}$$
$$= \begin{cases} (q_f(\exp(\gamma) + 1) + (q_c - q_f))\lambda, & x \leq q_t \\ (q_c(\exp(\gamma) + 1) - (q_c - q_f))\lambda, & x > q_t \end{cases}$$
$$= \begin{cases} (q_f(1/\lambda) + (q_c - q_f))\lambda, & x \leq q_t \\ (q_c(1/\lambda) - (q_c - q_f))\lambda, & x > q_t \end{cases}$$
$$= \begin{cases} q_f + (q_c - q_f)\lambda, & x \leq q_t \\ q_c - (q_c - q_f)\lambda, & x > q_t. \end{cases} \quad (5)$$

Note that $q_c - q_f = 1$, since $q_f$ and $q_c$ are obtained by applying $floor$ and $ceil$ functions, respectively, to the normalized input $x$. Using this fact,

$$\phi(x; \beta = \beta^*) = \begin{cases} q_f + \lambda, & x \leq q_t \\ q_c - \lambda, & x > q_t, \end{cases} \quad (6)$$

which corresponds to Eq. (9) in the main paper.

## 2. Overall Process of DAQ

We show in Algorithm. 1 an overall quantization process of DAQ.

---
*Corresponding author

1

**Algorithm 1** The distance-aware quantizer (DAQ)

1: **Input**:
   $\hat{x}$: an element of either full-precision weights or activations.
2: **Output**:
   $Q(\hat{x})$: a quantized value with a $b$-bit precision.
3: **Parameters**:
   $\gamma$: a positive constant, $l$: a lower bound, $u$: an upper bound.
4: **Training**:
5: Clip and normalize the input $\hat{x}$:
   $x = (2^b - 1)\{\text{clip}(\hat{x}, \min = l, \max = u) - l\}/(u - l)$.
6: Find the two nearest quantized values, $q_f$ and $q_c$:
   $q_f = \text{floor}(x), q_c = \text{ceil}(x)$.
7: Compute distance scores and adjust the temperature $\beta^*$:
   $d_x(q_i) = \exp(-|x - q_i|), i \in \{f, c\}$
   $\beta^* = \gamma/|s_x(q_f) - s_x(q_c)|$,
   where $s_x(q)$ is a weighted score, defined as $k_x(q)d_x(q)$.
8: Compute distance probabilities $m_x$ for quantized values, $q_f$ and $q_c$, with an adaptive temperature $\beta^*$:
   $m_x(q_i; \beta = \beta^*) = \frac{\exp(\beta^* s_x(q_i))}{\sum_{j \in \{f,c\}} \exp(\beta^* s_x(q_j))}, i \in \{f, c\}$
9: Compute the assignment using the *continuous* assignment function $\phi(x; \beta = \beta^*)$ with the corresponding temperature $\beta^*$:
   $\phi(x; \beta = \beta^*) = \sum_{i \in \{f,c\}} m_x(q_i; \beta = \beta^*)q_i$.
10: Rescale the assignment $\phi(x; \beta = \beta^*)$ using the following function:
    $f(y) = (y - q_t)/(1 - 2\lambda) + q_t$,
    where $\lambda = 1/(e^\gamma + 1)$ and $q_t = (q_f + q_c)/2$.
11: Set $Q(\hat{x}) = f(\phi(x; \beta = \beta^*))$.
12: **Inference**:
13: Clip and normalize the input $\hat{x}$:
    $x = (2^b - 1)\{\text{clip}(\hat{x}, \min = l, \max = u) - l\}/(u - l)$.
14: Find the nearest quantized value using a rounding function:
    $Q(\hat{x}) = \text{round}(x)$.

## 3. Design of DAQ

We design functions in DAQ for distance scores and kernels satisfying the following conditions. The distance score $d_x$ increases as the input $x$ approaches the quantized value $q$, and the kernel function $k_x$ retains values nearby its center. Under these conditions, we can use other functions for distance scores and kernels. To test this, we apply variants of them to our quantizer, where the parameters of $\gamma$ and standard deviation of the kernel are fixed to the values chosen from the original functions. Specifically, we quantize ResNet-20 in an 1/1-bit setting, and test the model on the test split of CIFAR-10. 1) For the distance score, we use two variants, $d_x^{v1}$ and $d_x^{v2}$, as follows:

$$d_x^{v1}(q) = 1 - |x - q|, \quad d_x^{v2}(q) = \frac{1}{1 + |x - q|}. \quad (7)$$

With these functions but using the kernel as our original Gaussian one, our quantizer achieves top-1 accuracies of 84.4 and 84.2, respectively. 2) For the kernel, we replace the Gaussian kernel with the Laplacian one while fixing the distance score as our original one in Eq. (4) of the main paper, achieving the top-1 accuracy of 83.7. These results suggest that switching the distance or kernel functions provides comparable results with the original one (85.8), even without tuning hyperparameters.

## 4. Experiments on the SR Task

We apply our method to the SR task on the Set5 [1] dataset. We quantize weights and activations for FSR-CNN [3] in 3/3, 4/4, and 8/8-bit settings, achiving the average peak signal to noise ratio (PSNR) of 35.99dB, 36.56dB, and 37.18dB, respectively, with the scale factor of 2, where the full-precision model shows the PSNR of 37.05dB. We can see the experimental results from the SR task show trends similar to those for the classification task in that the high-bit quantized model (*i.e.*, 8/8-bit setting) provides a better result than the full-precision one. This suggests that DAQ could be effective in the regression task as well. Note that most quantization methods [2, 4, 5, 6, 7, 8] show their effectivenesses only to the classification task, making it difficult to compare them directly with ours on the SR task.



\end{document}